\documentclass[letterpaper]{article}

\usepackage{natbib,alifeconf}  
\usepackage[ruled,vlined]{algorithm2e}
\usepackage{amsmath}
\usepackage{amssymb}

\title{DIAS: A Domain-Independent Alife-Based Problem-Solving System}
\author{Babak Hodjat$^1$, Hormoz Shahrzad$^{1,2}$, and Risto                        
  Miikkulainen$^{1,2}$\\$^1$Cognizant AI Labs; $^2$The University of Texas at   
  Austin}

\begin{document}
\renewcommand{\floatpagefraction}{0.99}
\renewcommand{\textfraction}{0.01}
\maketitle
\pagestyle{plain}

\begin{abstract}
A domain-independent problem-solving system based on principles of Artificial Life is introduced. In this system, DIAS, the input and output dimensions of the domain are laid out in a spatial medium. A population of actors, each seeing only part of this medium, solves problems collectively in it. The process is independent of the domain and can be implemented through different kinds of actors. Through a set of experiments on various problem domains, DIAS is shown able to solve problems with different dimensionality and complexity, to require no hyperparameter tuning for new problems, and to exhibit lifelong learning, i.e.\ adapt rapidly to run-time changes in the problem domain, and do it better than a standard non-collective approach. DIAS therefore demonstrates a role for Alife in building scalable, general, and adaptive problem-solving systems.
\end{abstract}

\section{Introduction}

Ecosystems in nature consist of diverse organisms each with a generic goal to survive. Survival may require different strategies and actions at different times. Emergent behavior from the collective actions of these organisms then makes it possible for the ecosystem as a whole to adapt to a changing world, i.e.\ solve new problems as they appear.

Such continual adaptation is often necessary for artificial agents in the real world as well. As a matter of fact, the field of reinforcement learning was initially motivated by such problems: The agent needs to learn while performing the task. While many offline extensions now exist, minimizing regret and finding solutions in one continuous run makes sense in many domains. For instance, there are domains where the fundamentals of the domain are subject to rapid and unexpected change, such as trading in the stock market, and control systems for functions that exhibit chaotic behavior.
Similarly in many game-playing domains opponents improve and change their strategies as they play. There are also domains where numerous similar problems need to be solved and there is little time to adapt to each one, such as trading systems with a changing portfolio of instruments, financial predictions for multiple businesses/units, optimizing multiple industrial production systems, optimizing growth recipes for multiple different plants, and optimizing designs of multiple websites.

However, current Artificial Intelligence (AI) systems are not adaptive in this manner. They are strongly tuned to each particular problem, and adapting to changes in it and to new problems requires much domain-specific tuning and tailoring.

The natural ecosystem approach suggests a possible solution: Separate the AI from the domain. A number of benefits could result: First, the AI may be improved in the abstract; it is possible to compare versions of it independently of domains. Second, the AI may more easily be designed to be robust against changes in the domain, or even switches between domains. Third, it may be designed to transfer knowledge from one domain to the next. Fourth, it may be easier to make it robust to noise, task variation, and unexpected effects, and to changes to the action space and state space.

This paper aims at designing such a problem-solving system and demonstrating its feasibility in a number of benchmark examples. In this Domain Independent Alife-based Problem Solving System (DIAS), a population of actors cooperate in a spatial medium to solve the current problem, and continue doing so over the span of several changing problems. The experiments will demonstrate that
\begin{itemize}
\itemsep -0.75ex
\item The behaviors of each actor are independent from the problem definition;
\item Solutions emerge continually from collective behavior of the actors;
\item The actor behavior and algorithms can be improved independently of the domains;
\item DIAS scales to problems with different dimensionality and complexity;
\item Very little or no hyperparameter tuning is required between problems;
\item DIAS can adapt to a changing problem domain, implementing lifelong learning; and
\item Collective problem-solving provides an advantage in scaling and adaptation.
\end{itemize}
DIAS can thus be seen as a promising starting point for scalable, general, and adaptive problem solving, based on principles of Artificial Life.

\section{Related Work}

In most population-based problem-solving approaches, such as Genetic Algorithms \citep[GA;][]{mitchell:gaintro96,eiben:book15}, Particle Swarm Optimization \citep{Sengupta2018, Rodriguez2004}, and Estimation of Distribution Algorithms \citep{krejca2020theory}, each population member is itself a candidate solution to the problem. In contrast in DIAS, the entire population together represents the solution.

Much recent work in Artificial Life concentrates on exploring how fundamentals of biological life, such as reproduction functions, hyper-structures, and higher order species, evolved \citep{gershenson2018self}.  However, some Alife work also focuses on potential robustness in problem solving \citep{Hodjat1994IntroducingAD}. For instance, in Robust First Computing as defined by \cite{ackley2014indefinitely}, there is no global synchronization, perfect reliability, free communication, or excess dimensionality. DIAS complies to these principles as well. While it does impose periodic boundary conditions, these boundaries can expand or retract depending on the dimensionality of the problem.

This approach is most closely related to Swarm Intelligence systems \citep{bansal2019evolutionary}, such as Ant Colony Optimization \citep{deng2019improved}. The main difference is that the problem domain is independent from the environment in which the actors survive, i.e.\ the ecosystem, and a common mapping is provided from the problem domain to the ecosystem. This approach allows for any change in the problem domain to be transparent to the DIAS process, which makes it possible to change and switch domains without reprogramming or restarting the actor population.

Several other differences from prior work result from this separation between actors and problem domains. First, the algorithms that the actors run can be selected and improved independently of the domain and need not be determined a priori. Second, the fitness function for the actors, as well as the mapping between the domain reward function and the actors' reward function, is predefined and standardized, and need not be modified to suit a given problem domain.  Third, the actors' state and action spaces are fixed regardless of the problem domain. Fourth, there is no enforced communication mechanism among the actors. While the actors do have the facility to communicate point-to-point and communication might emerge if needed, it is not a precondition to problem solving.

In terms of prior work in the broader field of Universal AI and Domain Independence \citep{Hutter00atheory}, most approaches are limited to search heuristics, such as extensions to the A* algorithm \citep{stern2019domain}. Such approaches still require domain knowledge such as the goal state, state transition operators, and costs. While efficient, these approaches lack robustness, and are designed to work on a single domain at a time. They do not do well if the domain changes during the optimization process. In the case of domain-independent planning systems \citep{della2009upmurphi}, the elaborate step of modeling the problem domain is still required. Depending on the manner by which such modeling is done, the system will have different performance. In this sense DIAS aims at more general domain-independent problem solving than prior approaches.

\section{Method}

A population of independent actors is set up with the goal of surviving in a common environment called a \emph{geo}.  The input and output dimensions of the domain are laid out across the geo. Each actor sees only part of the geo, which requires that they cooperate in discovering collective solutions. This design separates the problem-solving process from the domain, allowing different kinds of actors to implement it, and makes it scalable and general. The population adapts to new problems through evolutionary optimization, driven by credit assignment through a contribution measure.

\begin{figure}[t]
\begin{center}
\includegraphics[width=0.85\columnwidth]{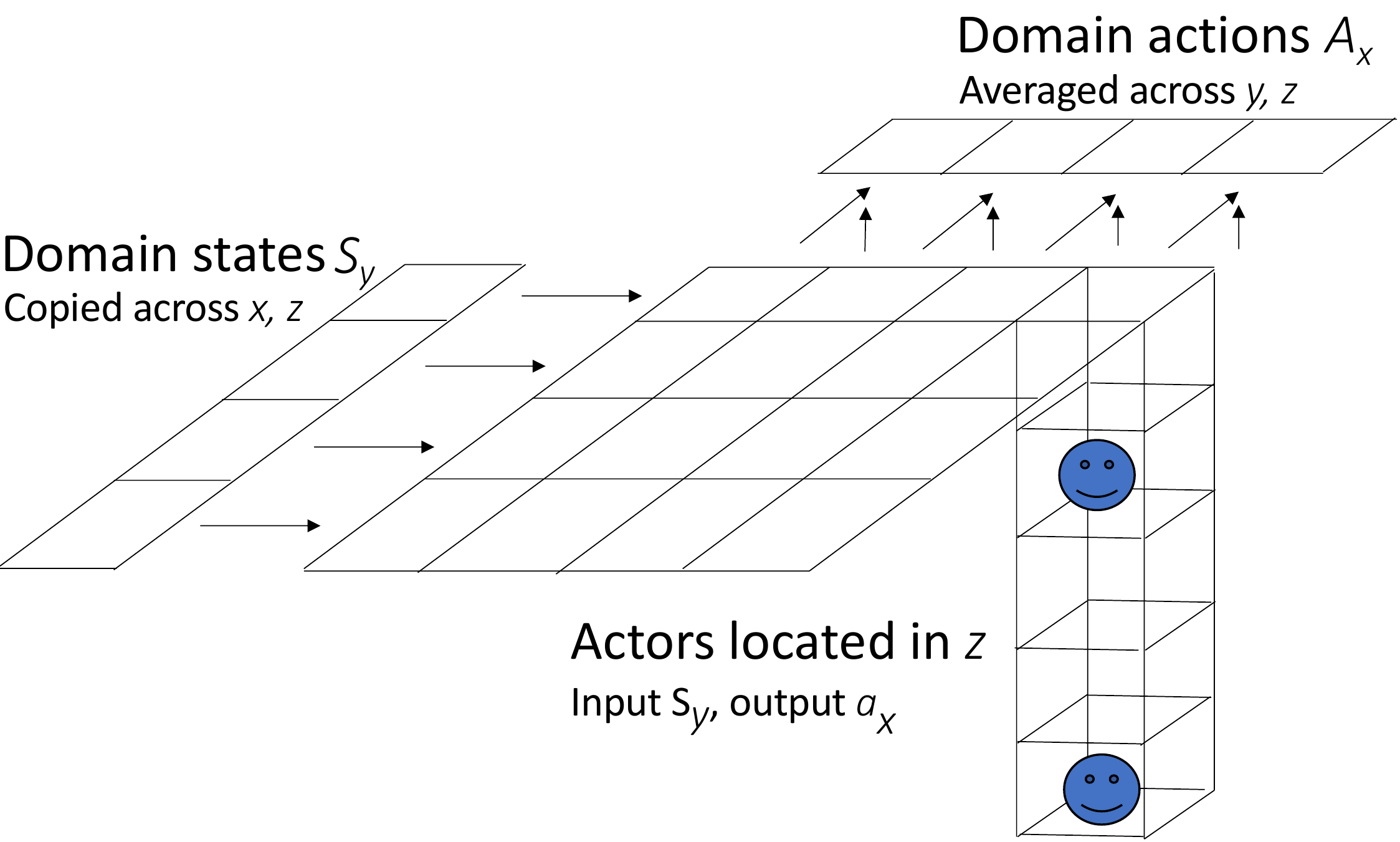}
\vspace*{-1ex}
\caption{General design of a DIAS system. Actors exist on a three-dimensional grid where $x$-locations represent the elements of the domain-action vector and $y$-locations represent the elements of the domain-state vector. The $z$-locations form a space that the actors can occupy and use for messaging. The grid thus maps the domain space to an actor space where problems can be solved in a domain-independent manner.\\[-3ex]}
\label{fig0}
\end{center}
\end{figure}

\subsection{Geo}

Actors are placed on a grid called geo (Fig.~\ref{fig0}). The dimensions of the grid correspond to the dimensions of the domain-action space (along the $x$-axis) and the domain-state space (along the $y$-axis). More specifically, domain action is a vector $\mathbf{A}$; each element $A_x$ of this vector is mapped to a different $x$-location. Similarly, domain state is a vector $\mathbf{S}$, and its elements $S_y$ are mapped to different $y$-locations in the geo. There can be multiple actors for each $(x, y)$-location of the grid. These actors live in different locations of the $z$ dimension. Each $(x, y, z)$ location may contain an actor, as well as a domain-action suggestion and a message, both of which can be overwritten by the actor in that location.

\subsection{Actors}

An actor is a decision-making unit taking an actor-state vector $\boldsymbol\sigma$ as its input and issuing an actor-action vector $\boldsymbol\alpha$ as its output at each domain time step. All actors operate in the same actor-state and actor-action spaces, regardless of the domain. Each actor is located in a particular $(x, y, z)$ location in the geo grid and can move to a geographically adjacent location. Each actor is also linked to a \emph{linked location} $(x', y', z')$ elsewhere in the geo. This link allows an actor to take into account relationships between two domain-action elements ($A_x$ and $A_{x'}$) and two domain-state elements ($S_y$ and $S_y'$) and to communicate with other actors via messages. Thus, it focuses on a part of the domain, and constitutes a part of a collective solution.

\vspace*{1ex}
The actor-action vectors $\boldsymbol\alpha$ consist of the following actions:
\begin{itemize}
\vspace*{-1.5ex}
\itemsep -0.75ex
\item Write a domain-action suggestion $a_x$ in the current location in the geo;
\item Write a message in the current location in the geo;
\item Write actor's reproduction eligibility;
\item Move to a geographically adjacent geo location;
\item Change the coordinates of the linked location.
\item NOP
\end{itemize}

The actor-state vectors $\boldsymbol\sigma$ consist of the following data:
\begin{itemize}
\vspace*{-1.5ex}
\itemsep -0.75ex
\item Energy $e$: real $\ge 0$;
\item Age: integer $\ge 0$;
\item Reproduction eligibility: True/False;
\item Coordinates in the current location: integer $x, y, z \ge 0$;
\item Message in the current location: [0..1];
\item Domain-action suggestion $a_x$ in current location: [0..1];
\item Domain-state value $S_y$ in the current location: [0..1];
\item Coordinates in the linked location: integer $x', y', z' \ge 0$;
\item Message in the linked location: [0..1];
\item Domain-action suggestion $a_{x'}$ in linked location: [0..1];
\item Domain-state value $S_{y'}$ in the linked location: [0..1].
\end{itemize}

Depending on the actor type, actors may choose to keep a history of actor states and refer to it in their decision making.

\subsection{Problem-solving Process}

Algorithm~\ref{alg:dias} outlines the DIAS problem-solving process. It proceeds through time intervals (in the main while loop). Each interval is one attempt to solve the problem, i.e.\ a fitness evaluation of the current system. Each attempt consists of a number of interactions with the domain (in the inner while loop) until the domain issues a terminate signal and returns a domain fitness. The credit for this fitness is assigned to individual actors and used to remove bad actors from the population and to create new ones through reproduction.

\begin{algorithm}[t]
\caption{The DIAS problem-solving process}
\label{alg:dias}
Initialize\_population; \emph{solved}=False; \emph{interval}=0\\
\While{interval $<$ maxinterval {\rm \&} $\neg$ solved}
{
1. Initialize\_domain; \emph{terminated}=False; $t$=0

2. \While{$t<$ maxt {\rm \&}  $\neg$ terminated}
  {
  2.1 Load $\mathbf{S}$\\
  2.2 \For{each actor}
      {
      input $\boldsymbol\sigma$\\
      output $\boldsymbol\alpha$
      }
  2.3 \For{each $x$}
      {
      Average $a_x$
      }
  2.4 Execute $\mathbf{A}$\\
  2.5 $t$++
  }

3. Obtain $F$

4. \If{$\neg$ solved}
  {
  4.1 \For{each actor}
      {
      Calculate $f$\\
      Calculate $\Delta e$\\
      \If{$e=0$}{Remove\_from\_population}
      }
  4.2 Reproduce\\
  4.3 \emph{interval}++
  }
}
\end{algorithm}

More specifically, during each domain time step $t$, the current domain-state vector $\mathbf{S}$ is first loaded into the geo (Step 2.1): Each $(x, y, z)$ location is updated with the domain-state element $S_y$. Each actor then takes its current actor state $\boldsymbol\sigma$ as input and issues an actor action $\boldsymbol\alpha$ as its output (Step 2.2). As a result of this process, some actors will write a domain-action suggestion $a_x$ in their location. A domain-action vector $\mathbf{A}$ is then created (Step 2.3): The suggestions $a_x$ are averaged across all locations with the same $x$ to form its elements $A_x$.  If no $a_x$ were written, $A_x(t-1)$ is used (with $A_x(-1)=0$). The resulting action vector $\mathbf{A}$ is passed to the domain, which executes it, resulting in a new domain state (Step 2.4).

Actors start the problem-solving process with an initial allotment of energy. After each interval (i.e.\ domain evaluation), this energy is updated based on how well the actor contributed to the performance of the system during the evaluation (Step 4.1). First, the domain fitness $F$ is converted into domain impact $M$, i.e.\ normalized within $[0..1]$ based on max and min fitness values observed in the past $R$ evaluations:
\begin{equation}
M = (F - F_{\min_R}) / (F_{\max_R} - F_{\min_R}).
\end{equation}
Thus, even though $F$ is likely to increase significantly during the problem-solving process, the entire range $[0..1]$ is utilized for $M$, making it easier to identify promising behavior.

Second, the contribution of the actor to $M$ is measured as the alignment of the actor's domain-action suggestions $a_x$ with the actual action elements $A_x$ issued to the domain during the entire time interval. In the current implementation, this contribution $c$ is 
\begin{equation}
\label{eq:contribution}
c = 1 - \min_{t=0..T} (|A_x(t) - a_x(t)|),
\end{equation}
where $T$ is the termination time; thus $c \in [0..1]$. The energy update $\Delta e$,  consists of a fixed cost $h$ and a reward that depends on the impact and the actor's contribution to it. If none of the actor's actions were 'write $a_x(t)$', i.e.\ the actor did not contribute to the impact,
\begin{equation}
\Delta e = h (M - 1),
\end{equation}
that is, the energy will decrease inversely proportional to impact. In contrast, if the actor issues one or more such 'write' actions during the interval,
\begin{equation}
\Delta e = h (cM (1-c) (1-M) - 1).
\end{equation}
In this case, the energy will also decrease (unless $M$ and $c$ are both either 0 or 1) but the relationship is more complex: It decreases less for actors that contribute to good outcomes (i.e.\ $M$ and $c$ are both high), and for actors that do not contribute to bad outcomes (i.e.\ the $M$ and $c$ are both low). Thus, regardless of outcomes, each actor receives proper credit for the impact. Overall, energy is a measure of the credit each actor deserves for both leading the system to success as well as keeping it away from failure. If an actor's energy drops to or below zero, the actor is removed from the geo.

For example, if the domain is a reinforcement learning game, like CartPole, each time interval consists of a number of left and right domain actions until the pole drops, or the time limit is reached (e.g.\ 200 domain time steps). At this point, the domain issues a termination signal, and the fitness $F$ is returned as the number of time steps the pole stayed up. That fitness is scaled to $M \in [0..1]$ using the max and min $F$ during the $R=60,000$ previous attempts. If $M$ is high, actors that wrote $a_x$ values consistently with $A_x$, i.e.\ suggested left or right at least once when those actions were actually issued to the domain, have a high contribution $c$, and therefore a small decrease $\Delta e$. Similarly, if the system did not perform well, actors that suggested left(right) when the system issued right(left), have a low contribution $c$ and receive a small decrease $\Delta e$. Otherwise the $\Delta e$ is large; such actors lose energy fast and are soon eliminated.

After each time interval, a number of new actors are generated through reproduction (Step 4.2). Two parents are selected from the existing population within each $(x, y)$ column, assuming the total energy in the column is below a threshold $E_{\max}$. If it is not, the agents are already very good, and evolution focuses on columns elsewhere where progress can still be made, or alternative solutions can be found.  In addition, a parent actor needs to meet a maturity age requirement, i.e.\ it must have been in the system for more than $V$ time intervals and not reproduced for $V$ time intervals. The actor also needs to have reproduction eligibility in its state set to True. 

Provided all the above conditions are met, a proportionate selection process is carried out based on actor fitness $f$, calculated as follows. First, the impact variable $M$ is discretized into $L$ levels: $M=\{b_0,b_1..,b_{L-1}\}$. Then, for each of these levels $b_i$, the probability $p_{i}$ that the actor's action suggestions align with the actual actions when $M=b_i$ is estimated as
\begin{equation}
p_i = P(c=1|M=b_i),
\end{equation}
where $c$ measures this alignment according to Eq.~\ref{eq:contribution}.
The same window of $R$ past intervals is used for this estimation as for determining the max and min $M$ for scaling the impact values. Finally, actor fitness $f$ is calculated as alignment-weighted average of the different impact levels $b_i$:
\begin{equation}
f = \sum_{i=0}^{L} p_i b_i.
\end{equation}
Thus, $f$ is the assignment of credit for $M$ to individual actors. Note that while energy measures consistent performance, actor fitness measures average performance. Energy is thus most useful in discarding actors and actor fitness in selecting parents.

Once the parents are selected, crossover and mutation are used to generate offspring actors. What is crossed over and mutated depends on the encoding of the actor type; regardless, each offspring's behavior, as well as its linked-location coordinates, is a result of crossover and mutation. Each pair of parents generates two offspring, whose location is determined randomly in the same $(x, y)$ column as the parents.

Note that the parents are not removed from the population during reproduction, but instead, energy is used as basis for removal. In this manner, the population can shrink and grow, which is useful for lifelong learning. It allows reproduction to focus on solving the current problem, while removal retains individuals that are useful in the long term. Such populations can better adapt to new problems and re-adapt to old ones.

Energy, age, and actor fitness for all actors in an $(x, y)$ column need to be available before reproduction can be done, so computations within the column must be synchronized in Step 4.2. However, the system is otherwise asynchronous across the $x$ and $y$ dimensions, making it possible to parallelize the computations in Steps 2 and 4. Thereby, the system scales to high-dimensional domains in constant time.

\subsection{Actor Types}

The current version of DIAS employs five different actor types:
\begin{itemize}
\itemsep -0.75ex
\item Random: Selects its next action randomly, providing a baseline for the comparisons;
\item Robot: Selects its next action based on human-defined preprogrammed rules designed for specific problem domains, providing a performance ceiling;
\item Bandit: Selects its next action using a basic multi-armed bandit algorithm (not including $\boldsymbol\sigma$ as context);
\item Q-Learning: Learns to select its next action using temporal differences; and
\item Rule-set Evolution: Evolves a set of rules to select its next action.
\end{itemize}

Simple Q-learning \citep{watkins1992q} was implemented based on the actor's state/action history, with the actor's energy difference from the prior time interval taken as the reward for the current interval. Because the dimensionality of the state/action space is limited by design, this method is a possible reinforcement learning approach.

Rule-set evolution \citep{Hodjat2018} was implemented based on rule sets that consist of a default rule and at least one conditioned rule. Each conditioned rule consists of a conjunction of one or more conditions, and an action that is returned if the conditions are satisfied. Conditions consist of a first and second term being compared, each with a coefficient that is evolved. An argument is also evolved for the action. Evolution selects the terms in the conditions from the actor-state space, and the action from the actor-action space. Rules are evaluated in order, and shortcut upon reaching the first to be satisfied. If none of the rules are satisfied, the default action is returned.

These actor types were evaluated in several standard benchmarks tasks experimentally, as will be described next.

\section{Experiments}

DIAS was evaluated in a number of benchmark problems to demonstrate the unique aspects of the approach. The system was shown scalable, general, and adaptable.  The dynamics of the problem-solving process were characterized and shown to be the source of these abilities.

\subsection{Test Domains}

In the n-XOR domain, the outputs of $n$ independent XOR gates, each receiving their own input, need to be predicted simultaneously. In order to make the domain a realistic proxy for real-world problems, 10\% noise is added to the XOR outputs. While a single XOR (or 1-XOR) problem can be solved by a single actor, solving $n > 1$ of them simultaneously requires a division of labor over the population. The different XOR input elements are in different $y$-locations and the different predicted outputs in different $x$-locations. With $n>1$, no actor can see or act upon the entire problem. Instead, emergent coordination is required to find behaviors that collectively solve all XORs. Increasing $n$ makes the problems exponentially more difficult (i.e.\ the chance of solving all $n$ XORs by luck is reduced exponentially with $n$).

The first set of experiments in the n-XOR domain show that the DIAS design scales to problems of different dimensionality and complexity.  The second set shows that DIAS can adapt to the different n-XOR problems online, i.e.\ to exhibit lifelong learning.

Experiments were also run on a number of OpenAI Gym games, including CartPole, MountainCar, Acrobot, and LunarLander. The same experimental setup was used across all of them without any hyperparameter tuning. The OpenAI Gym domains thus show that DIAS is a general problem-solving approach, requiring little or no parameter tuning when applied to new problems.

\begin{figure*}[t]
\begin{center}
\includegraphics[width=1.7in]{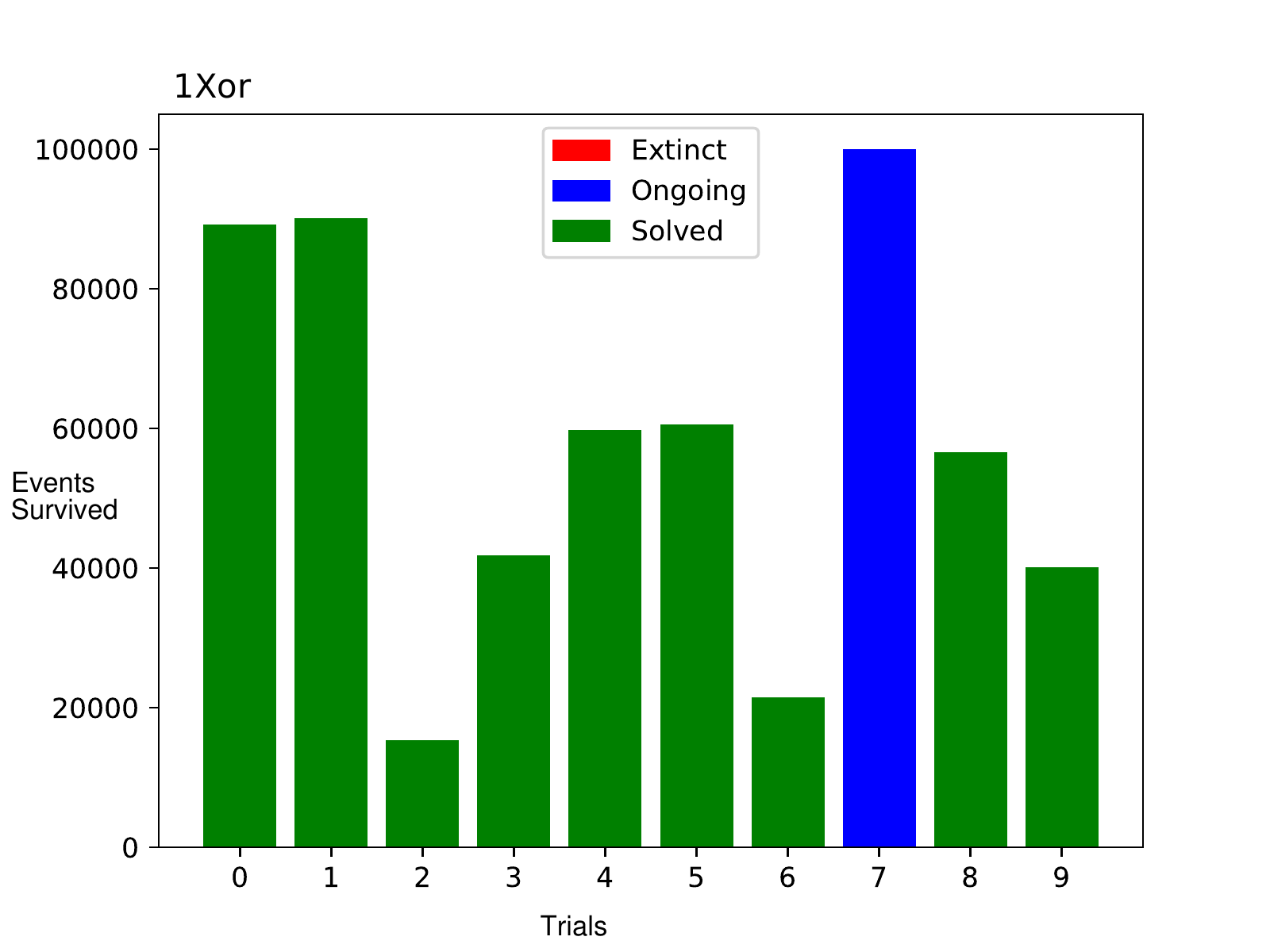}
\includegraphics[width=1.7in]{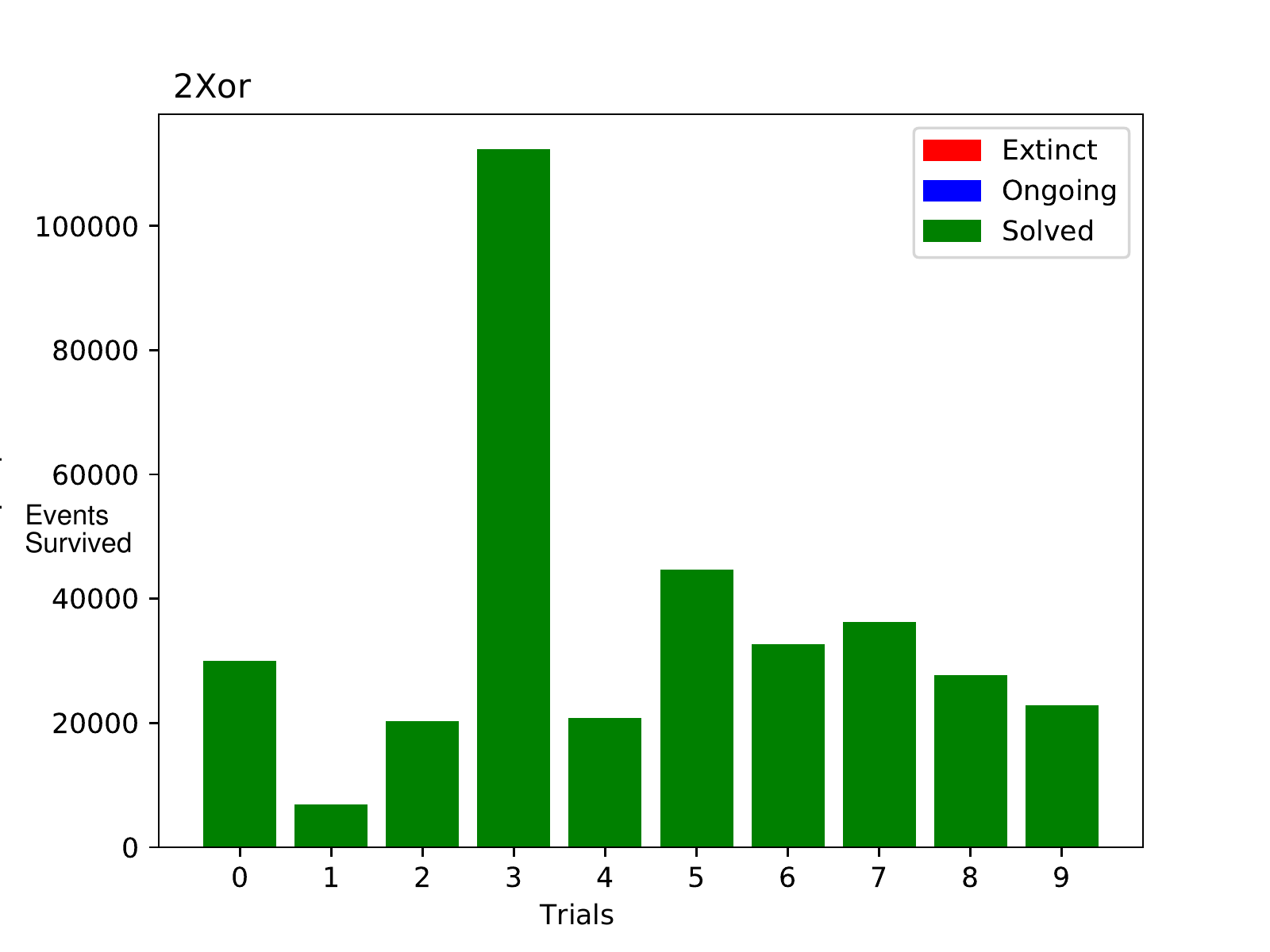}
\includegraphics[width=1.7in]{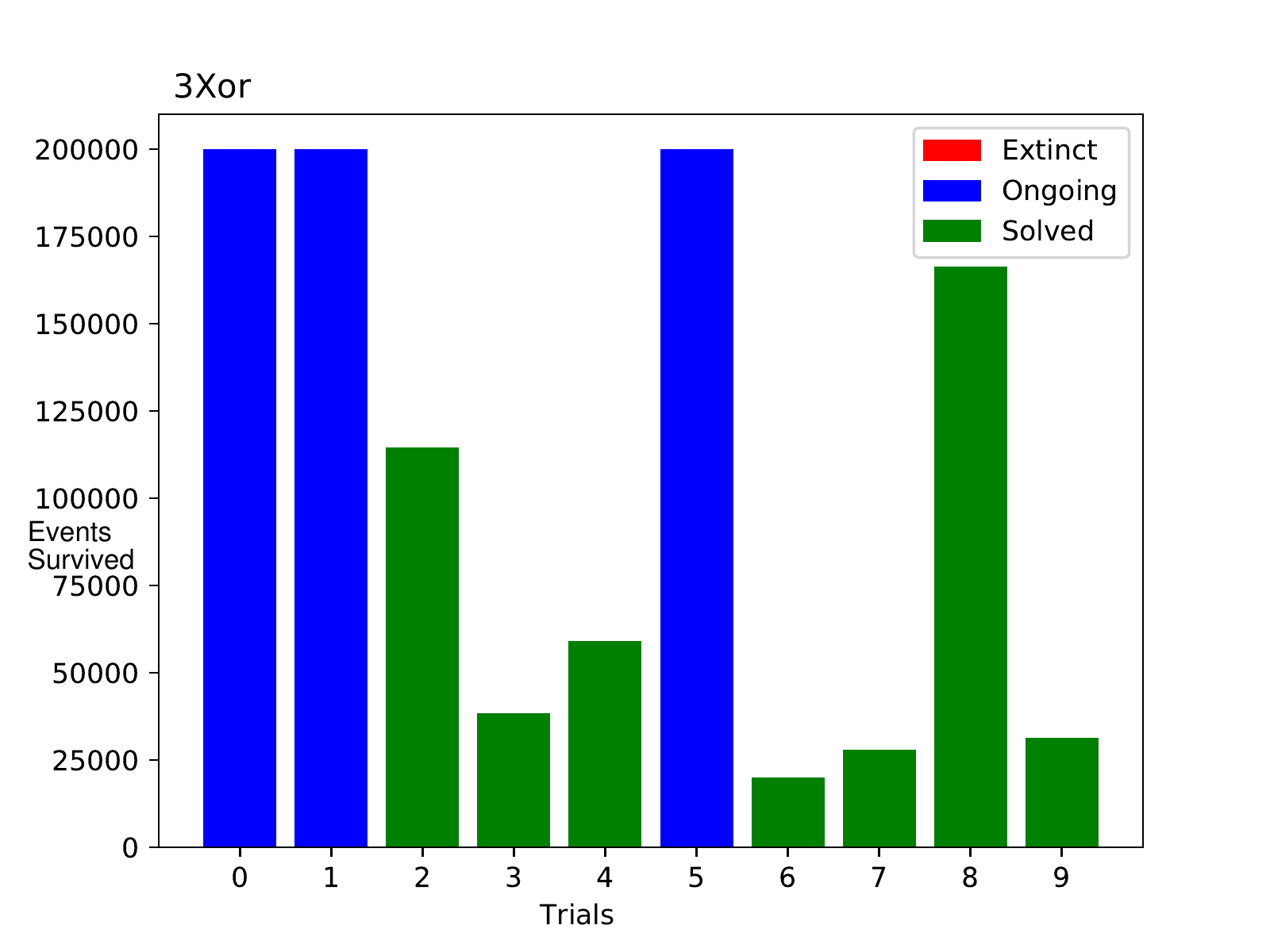}
\vspace*{-2ex}
\caption{Number of time intervals needed to solve the 1-XOR (left), 2-XOR (middle) and 3-XOR (right) problems in 10 independent runs. No runs lead to extinction (they would have been shown with red bars), though some do not completely solve the problem within the allotted 200,000 time intervals (these runs are shown with blue bars). These experiments show that the DIAS framework scales naturally to problems with increasing dimensionality and complexity.\\[-4ex]}
\label{fig1}
\end{center}
\end{figure*}

\begin{figure}[t!]
\scriptsize
\vspace*{1ex}
\begin{verbatim}
ancestor_count=1065
action counts={decrease_linked_location_y: 55, write: 19}
total_potential_contribution_count=19
total_contribution_count=19
impact_contribution_probabilities=1.0:1.0
state={energy: 66.0; age: 74;
       reproduction_eligibility: True;
       own_location_coordinates 0, 1, 18;
       own_location_message: 0;
       own_location_domain_action: None;
       own_location_domain_state: 0;
       linked_location_coordinates: 0, 1, 19;
       linked_location_message: 0;
       linked_location_domain_action: None;
       linked_location_domain_state: 0;
Rule1<49>: (0.21*y <= 0.62*linked_location_domain_state)
            --> decrease_linked_location_y(0.10)
Rule2<0>:  (0.9*own_location_domain_state 
             > 0.15*reproduction_eligibility)&
           (0.21*y <= 0.62*linked_location_domain_state)&
           (0.21*y <  0.62*linked_location_domain_state)
           --> decrease_linked_location_y(0.10)
Rule3<6>:  (0.90*own_location_domain_state
             > 0.15*reproduction_eligibility)
           --> decrease_linked_location_y(0.10)
Rule4<0>:  (0.21*y < 0.62*linked_location_domain_state)
           --> decrease_linked_location_y(0.10)
Deflt<19>: --> write(0.93)
\end{verbatim}
\vspace*{-3ex}
\caption{An example actor that solves the 1-XOR problem, consisting of a number of metrics, current state, and a set of rules. 
The 'write' action writes its argument in the own\_location\_domain\_action field as the actor's suggested domain action $a_x$. Even though the rules explicitly describe the actor's behavior, it is not possible to tell from this one actor what the solution to the complete problem is. The actor does not see the whole problem or determine the outcome alone: The population as a whole collectively solves it.\\[-4ex]}
\label{fig:xoractor}
\end{figure}

\subsection{Experimental Setup}

Each experiment consists of 10 independent runs of up to 200,000 time intervals. For each domain, the number of $x$-locations is set to the number of domain actions, and the number of $y$-locations to the number of domain states (1, 2 for 1-XOR; 2, 4 for 2-XOR; 3, 6 for 3-XOR; 2, 4 for CartPole; 3, 2 for MountainCar; 3, 6 for Acrobot; and 3, 6 for LunarLander). The number of $z$-locations is constant at 100 in all experiments. The initial population for each $(x, y)$ location is set to 20 actors, placed randomly in $z$. Each Q-learning actor is initialized with random Q-values, and each rule-set actor with a random default rule. The robot and bandit actors have no random parameters, i.e.\ they are all identical. 

The range $R$ used for scaling domain fitnesses to impact values was 60,000 intervals, and the impact $M$ was discretized into 21 levels $\{0, 0.05, .., 0.95, 1\}$ in calculating actor fitness. Each actor started with an initial energy of 100 units, with a fixed cost $h=2$ units at each time interval. The energy threshold $E_{\max}$ for reproduction in each $(x, y)$ column was set to the initial energy, i.e.\ 20 * 100 = 2000 (note that while each actor's energy decreases over time, population growth can increase total energy). Reproduction eligibility was set to True at birth, and the reproduction maturity requirement $V$ to 20. Small variations to this setup lead to similar results. In contrast, each of the main design choices of DIAS is important for its performance, as verified in extensive preliminary experiments.

Each experiment can result in one of three end states: (1) the actor population solves the problem; (2) all actors run out of energy before solving the problem and the actor population goes extinct; and (3) the actor population survives but has not solved the problem within the maximum number of time intervals. In practice, it is possible to restart the population if it goes extinct or does not make progress in $F$ after a certain period of time. Restarts were not implemented in the experiments in order to evaluate performance more clearly.

For comparison, direct evolution of rule sets (DE) was also implemented in the DIAS framework. The setup is otherwise identical, but a DE actor receives the entire domain state vector $\mathbf{S}$ as its input and generates the entire domain action vector $\mathbf{A}$ as its output. DE therefore does not take advantage of collective problem solving. A population of 100 DE actors is evolved for up to 100,000 time intervals through a GA with $F$ as the individual fitness, tournament selection, 25\% elitism, and the same crossover and mutation operators as in DIAS.

\subsection{Comparing Actor Types}

The five actor types described above were each tested in preliminary experiments on 1-XOR, using the same settings. These results demonstrate that collective behavior resulting from the DIAS framework can successfully solve these domains. 

The Robot actor specifically written for 1-XOR solves it from the first time interval. Similarly, a custom-designed Robot actor is always successful in the CartPole domain. On the other hand, Random, Bandit, and Simple Q-Learning were not able to solve 1-XOR at all: Each attempt leads to extinction in under 350 time intervals. While it is possible that these actors could solve simpler problems, the search space for 1-XOR is apparently already too large for them. Therefore, the main experiments focus on the Rule-set Evolution actor type. 

\begin{figure*}[t]
\begin{center}
\includegraphics[width=1.7in]{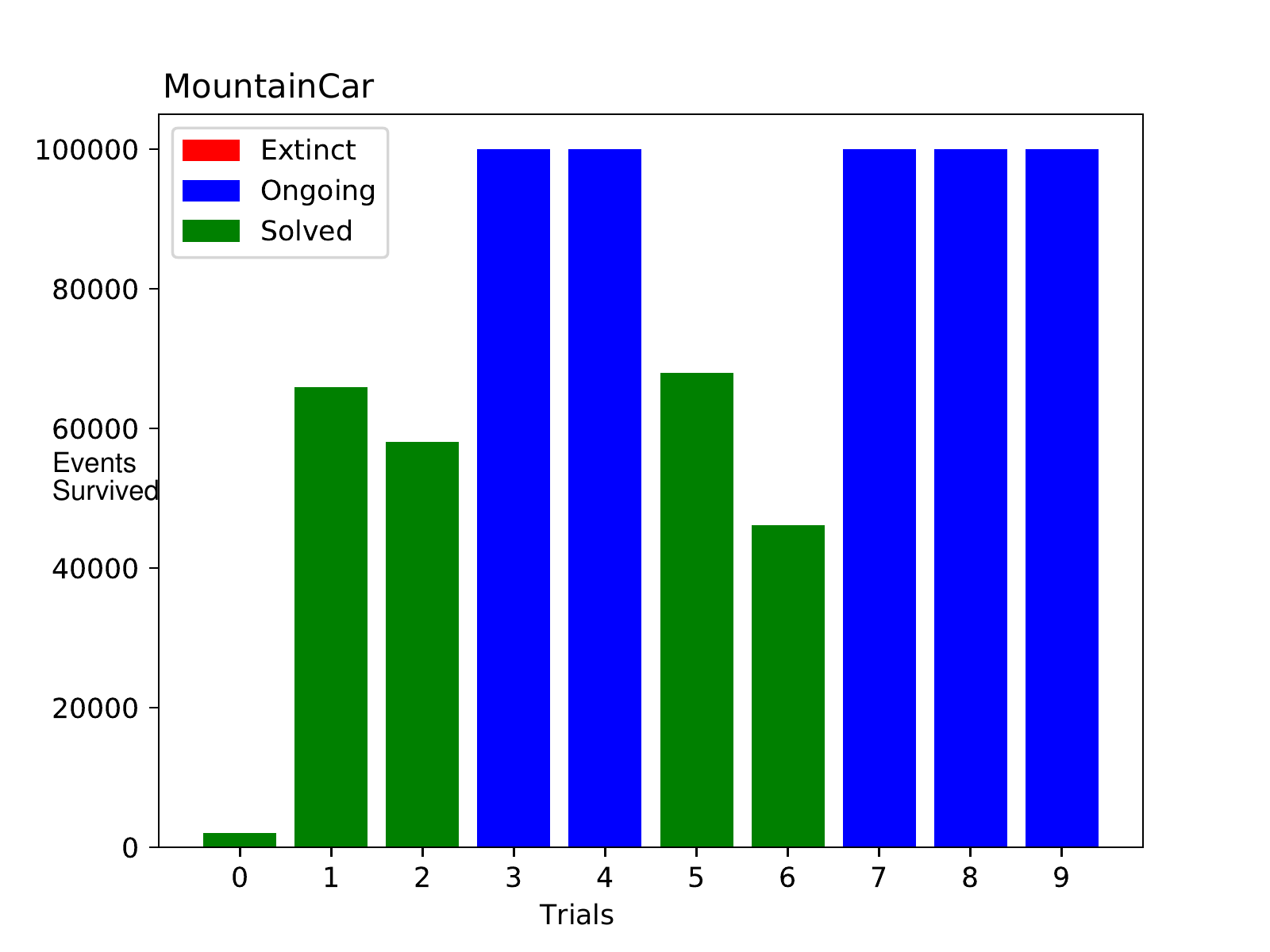}
\includegraphics[width=1.7in]{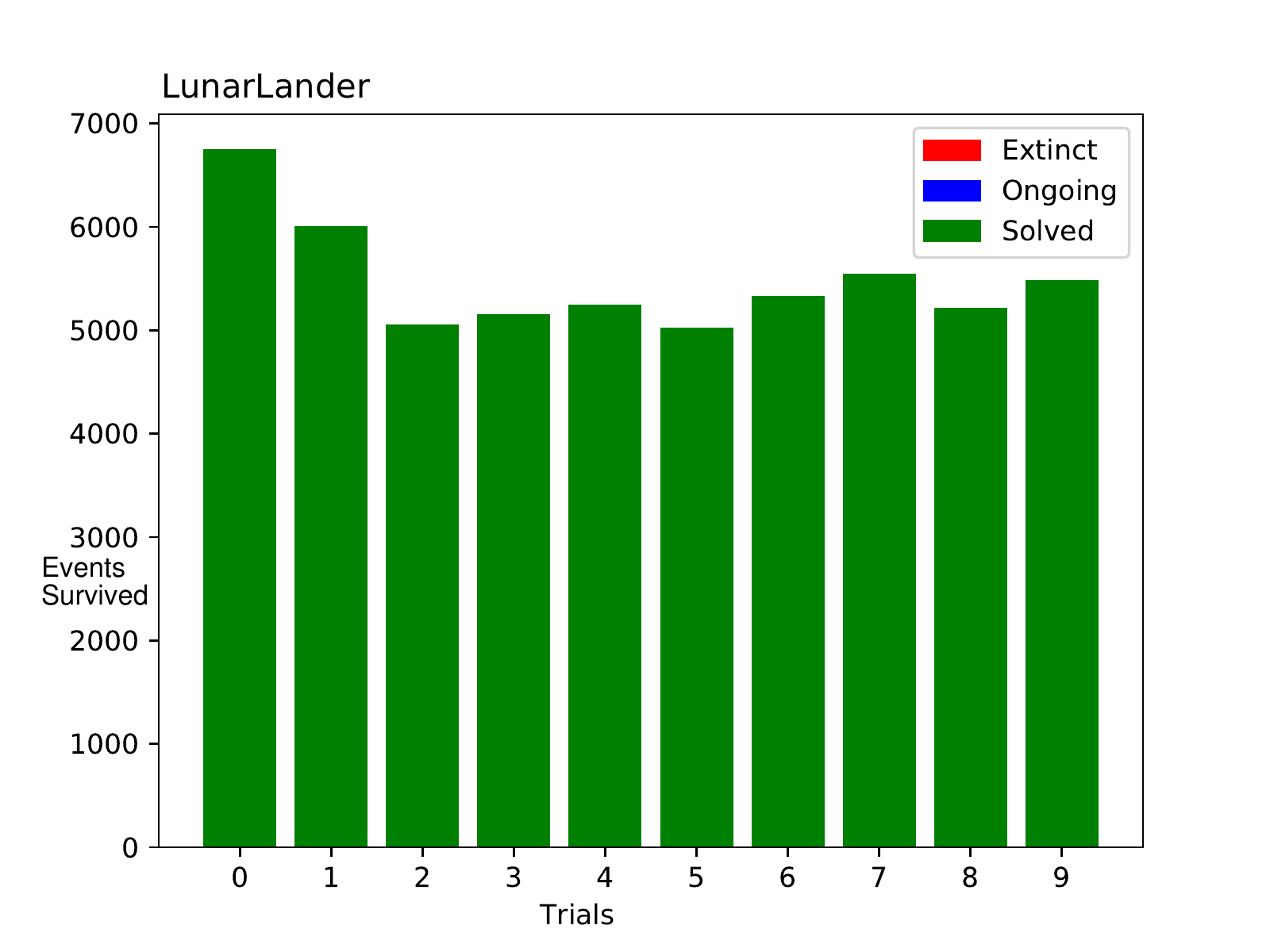}
\includegraphics[width=1.7in]{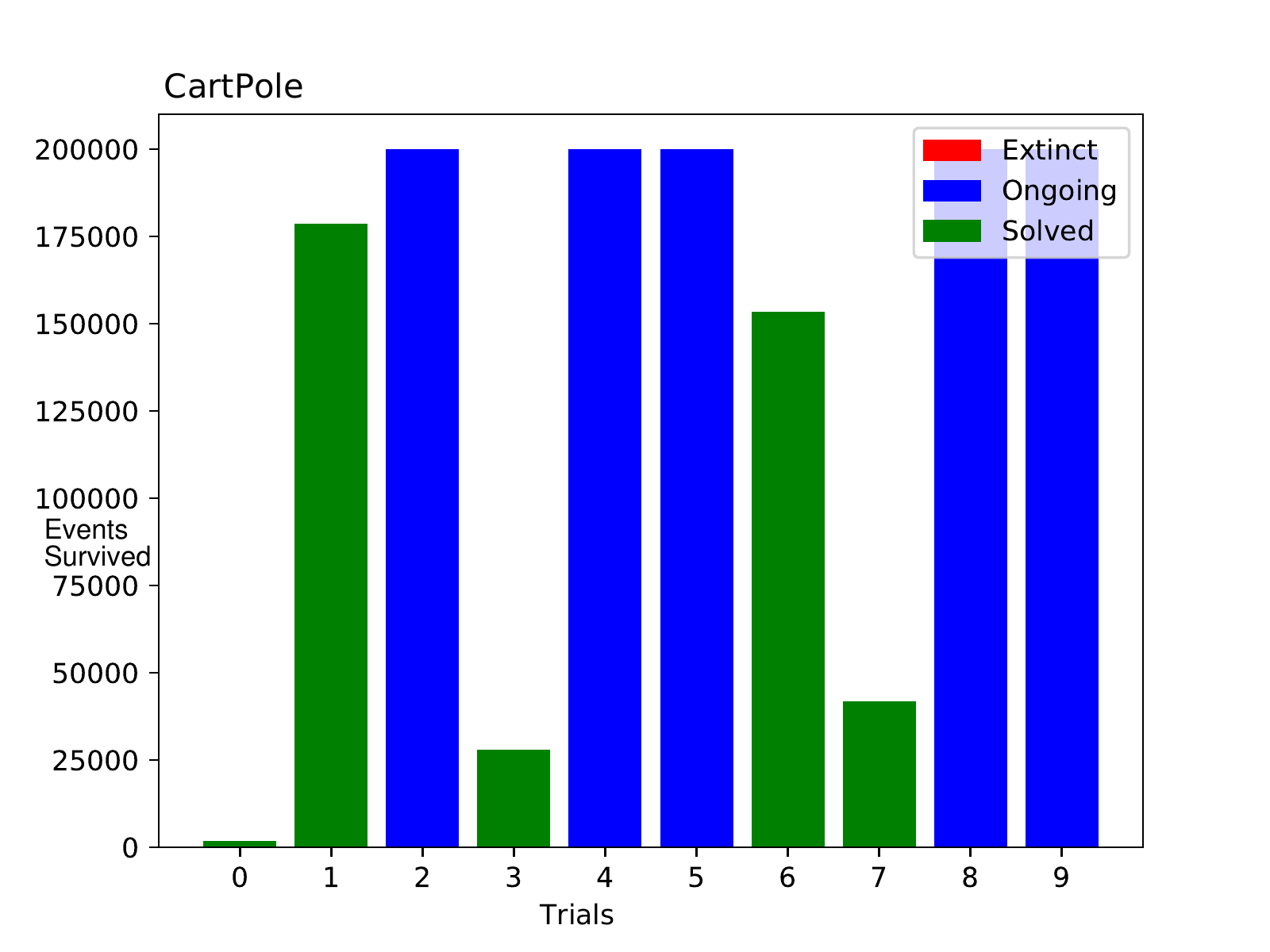}
\includegraphics[width=1.7in]{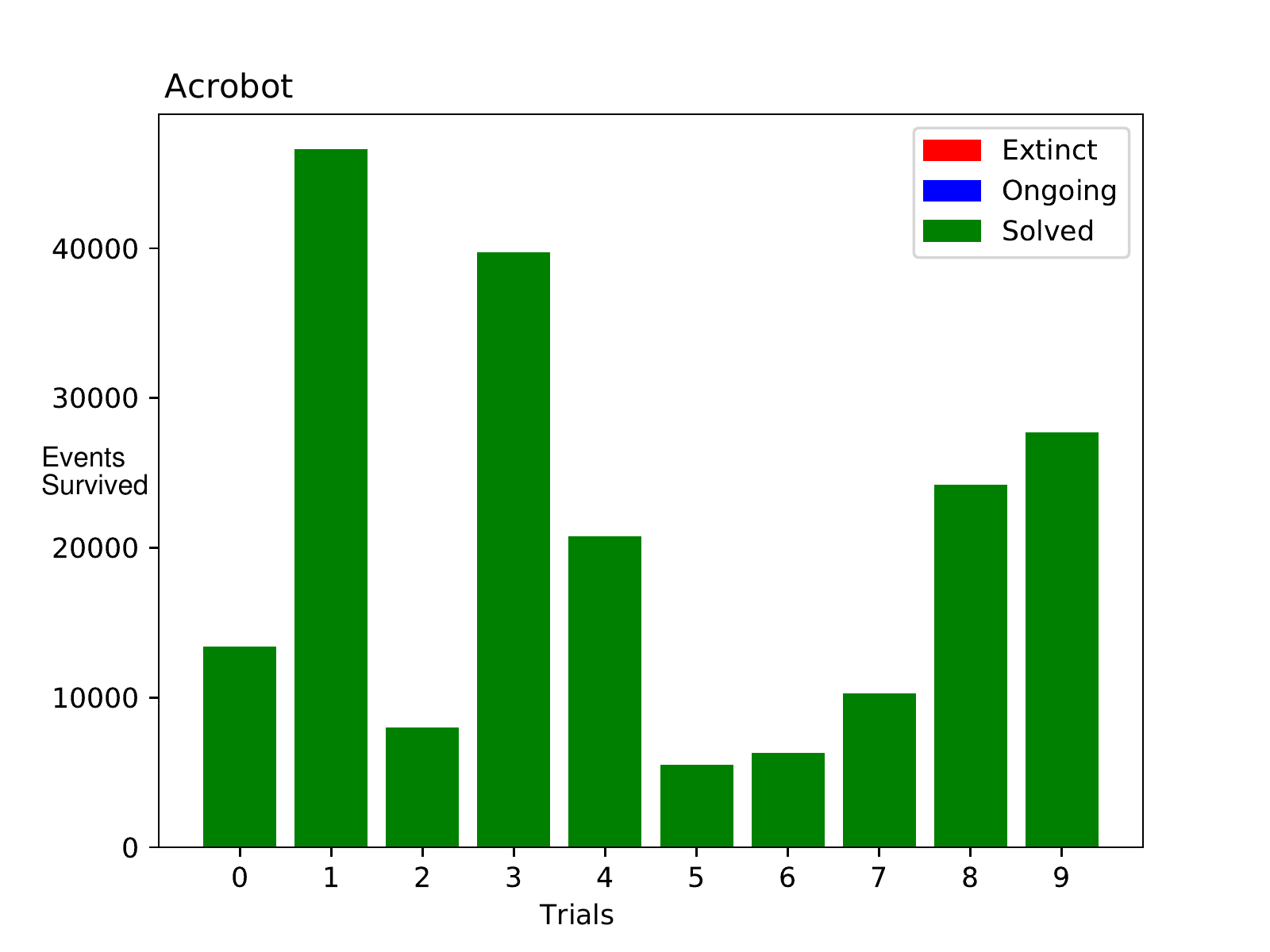}
\vspace*{-2ex}
\caption{Solving different kinds of problems in the OpenAIGym domain. Results of 10 runs in the MountainCar (left), LunarLander (second from left), CartPole (second from right), and Acrobot (right) problems are shown. Again, no runs resulted in extinction, although some MountainCar and Cartpole runs did not completely solve the problem within the allotted maximum number of time intervals. Notably, DIAS solves all these problems, as well as all other domains in the paper, with the same hyperparameter and experimental settings, demonstrating the generality of the approach.\\[-4ex]}
\label{fig4}
\end{center}
\end{figure*}

\begin{figure}[t]
\begin{center}
\vspace*{1ex}
\includegraphics[width=\columnwidth]{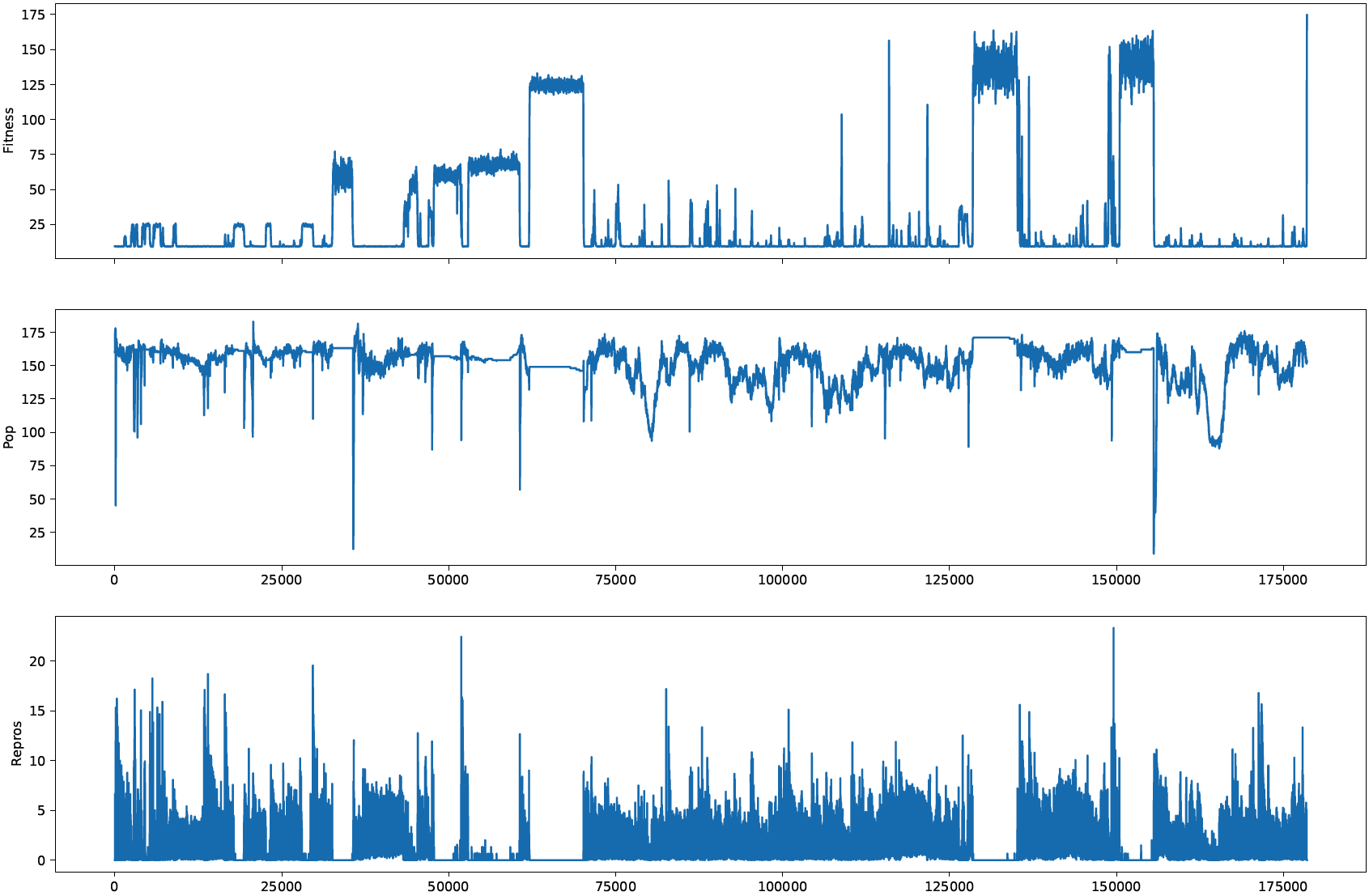}
\vspace*{-4ex}
\caption{Population dynamics in a sample run of the CartPole problem, showing progression of domain fitness $F$ (the number of time steps the pole stays upright; top), number of live actors (middle), and the number of reproductions (bottom) at each time interval. The reproductions drop and the population becomes relatively stable during periods when the ecosystem finds a peak in $F$; however, these peaks are unstable and the population eventually moves on to explore other solutions. Such dynamics make it possible to not only find solutions to the current problem, but to also adapt rapidly to changing domains and new problems.\\[-6ex]}
\end{center}
\label{fig5}
\end{figure}


\subsection{Scaling to Problems of Varying Complexity}

The first set of main experiments showed that the DIAS population solves n-XOR with
$n=$1, 2, and 3 reliably (Fig.~\ref{fig1}).  Even with 10 percent reward noise, the system is resilient and the population collectively achieves the best possible reward, even if it is not constant over time.  In comparison, while DE solved the 1-XOR in less than 10,000 time intervals in nine of 10 runs, only three runs solved the 2-XOR and none solved the 3-XOR within 100,000 time intervals. These results show that DIAS provides an advantage in scaling to problems with higher dimensionality and complexity.

The success was due to emergent collaborative behavior of the actor population. This result can be seen by analyzing the rule sets that evolved, for example that of the actor from a population that solved the 1-XOR problem, shown in Fig.~\ref{fig:xoractor}. This actor is number 1065 in its lineage. It has contributed to the domain action 19 times, and all 19 times, its contribution has been in line with the domain action issued. Therefore, the vector of alignment probabilities $p_{i}$ at each impact level $i$ has only one element: The probability is 1.0 for the impact level of 1.0. Its current state is high in energy for its age, suggesting that it has contributed well. Its current linked location has null values in message, domain-action, and domain-state fields.

In terms of rules, the second and fourth are redundant, and never fired (redundancy is common in evolution because it makes the search more robust). Rule~1 fired 49 times, Rule~3 six times, and the default rule 19 times. Rules~1 and~3 perform a search for a linked location that has a large enough domain-state value: They decrease the $y$-coordinate of the linked location whenever they fire. If such a location is found (Rule~1), and its own domain-state value is high enough (Rule~3), 0.93 is written as its suggested domain action $a_x$ (Default rule). An $a_x > 0.5$ denotes a prediction that the XOR output is~1, while $a_x\le 0.5$ suggests that it is~0; therefore, this actor contributes to predicting XOR output 1.\ Other actors are required to generate the proper domain actions in other cases. Thus, problem solving is collective: Several actors need to perform compatible subtasks in order to form the whole solution.

\subsection{Solving Different Kinds of Problems}

The second set of main experiments was designed to demonstrate the generality of DIAS, i.e.\ that it can solve a number of different problems out of the box, with no change to its settings. CartPole, MountainCar, Acrobot, and LunarLander of OpenAI Gym were used in this role because they represent a variety of well-known reinforcement-learning problems.

DIAS was indeed able to solve each of these problems without any customization, and with the same settings as the n-XOR problems (Fig.~\ref{fig4}).  A histogram of the population dynamics as the ecosystem evolves to a solution is shown in Fig.~\ref{fig5} for the CartPole problem. The system gradually finds higher domain fitness peaks, and every time it does so, the number of reproductions drop and the population stabilizes.  In this manner, DIAS is trying out different equilibria, eventually finding one that implements the best solution.  


\subsection{Adapting to Changing Problems}

A third set of experiments were run in the n-XOR domain to demonstrate the system's ability to switch between domains mid-run. The run starts by solving the 1-XOR problem; then the problem switches to 2-XOR, 3-XOR, and back to 1-XOR again. Note that the max domain fitness level also changes mid-run as problems are switched. These switches require the geo to expand and retract, as the dimension of $x$ (i.e.\ number of domain actions) and $y$ (number of domain states) are different between problems. This change, however, does not affect the actors, whose action and state spaces remain the same. When retracting, actors in locations that no longer exist are removed from the system. When expanding, new actors are created in locations $(i, j, k)$ with $i>x$ and/or $j>y$ by duplicating the actor in location $(i~\mathrm{mod}~x, j~\mathrm{mod}~y, k)$, if any.

\begin{figure*}[t]
\centering
\hfill
\includegraphics[height=2.5in]{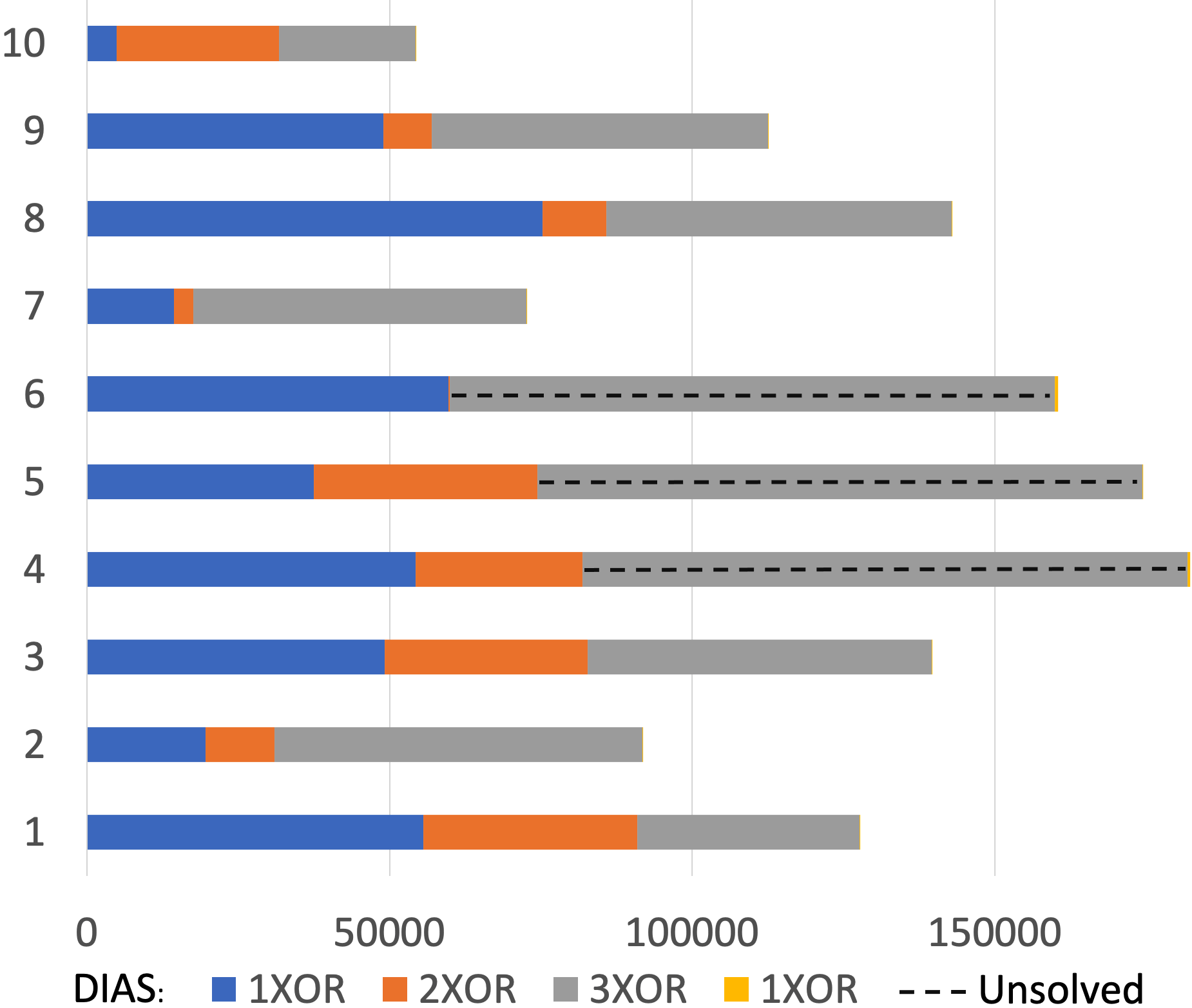}
\hfill
\includegraphics[height=2.5in]{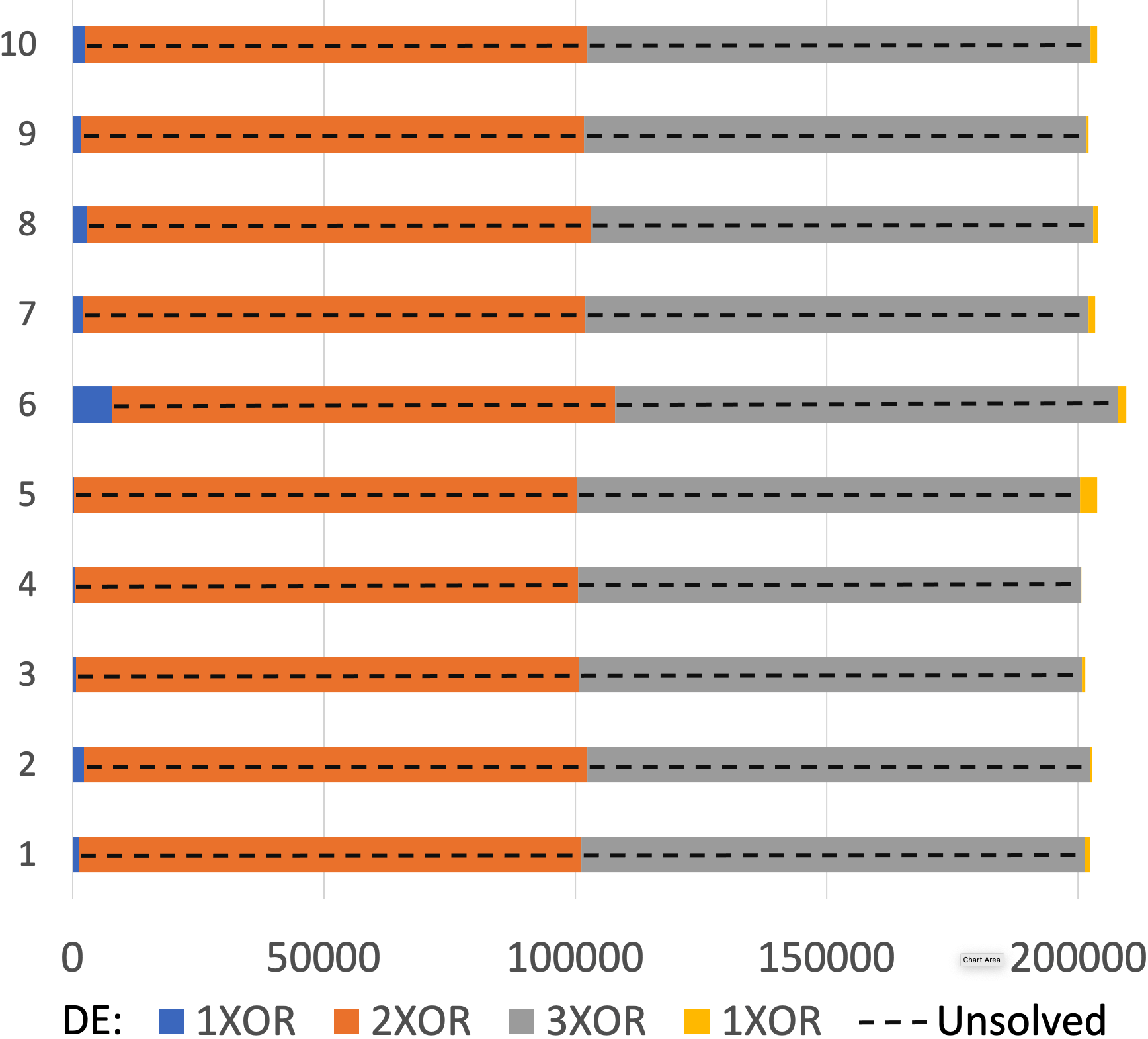}
\hfill\mbox{}
\caption{\label{fig:switch}Adapting to changing problems. Ten runs of DIAS (left) and DE (right) are shown where the problem switched from 1-XOR to 2-XOR, 3-XOR, and back to the 1-XOR as soon as the problem was solved or 100,000 time intervals passed (dashed line). DIAS was able to adapt to new problems quickly, solve new problems quicker, and particularly quickly when returning to 1-XOR. In contrast, while DE solved the first 1-XOR quickly, it was not able to adapt to 2-XOR nor 3-XOR mid-run, and it did not solve the second 1-XOR faster than the first. Thus, collective problem solving in DIAS provides a significant advantage in adapting to new problems, i.e.\ in lifelong learning.}
\end{figure*}

The results of 10 such runs are shown in Fig.~\ref{fig:switch}. In seven of these runs, DIAS was able to solve the entire sequence of problems. Most interestingly, the time it needed for subsequent problems often became shorter. For example Run~1 
took 55,574 time intervals to solve the 1-XOR problem, another 35,363 to solve the 2-XOR, and 36,690 more to solve the 3-XOR. Then, switching back to the 1-XOR problem, a solution was found within a mere 51 time intervals. These results demonstrate that DIAS is able to adapt to new problems quickly, retain information from earlier problems, and utilize it in later problems.

In contrast, while DE solved the 1-XOR fast in the beginning and end of each sequence, none of its 10 runs were able to adapt to 2-XOR and 3-XOR mid-run. Also, it did not solve the second 1-XOR any faster than the first one.

These experiments thus show that the collective problem solving in DIAS is essential for solving new problems continuously as they appear, and for retaining the ability to solve earlier problems. In this sense, it demonstrates an essential ability for continual, or lifelong, learning. It also demonstrates the potential for curriculum learning for more complex problems: The same population can be set to solve domains that get more complex with time. Such an approach may have a better chance of solving the most complex problems than one where they are tackled directly from the beginning.

\section{Discussion and Future Work}

The experimental results with DIAS are promising: They demonstrate that the same system, with no hyperparameter tuning or domain-dependent tweaks, can solve a variety of domains, ranging from classification to reinforcement learning. The results also demonstrate ability to switch domains in the middle of the problem-solving process, and potential benefits of doing so as part of curriculum learning. The system is robust to noise, as well as changes to its domain-action space and domain-state space mid-run.

The most important contribution of this work is the introduction of a common mapping between a domain and an ecosystem of actors. This mapping includes a translation of the state and action spaces, as well as a translation of domain rewards to the actors contributing (or not contributing) to a solution. It is this mapping that makes collective problem solving effective in DIAS. With this mapping, changes to the domain have no effect on the survival task that the actors in the ecosystem are solving. As a result, the same DIAS system can solve problems of varying dimensionality and complexity, solve different kinds of problems, and solve new problems as they appear, and do it better than DE can.

In this process, interesting collective behavior analogous to biological ecosystems can be observed. Most problems are being solved through emergent cooperation among actors (i.e.\ when $x$ and/or $y$-dimensionality $> 1$). Problem solving is also continuous: The system regulates its population, stabilizing it as better solutions are found. Because of this cooperative and continual adaptation, it is difficult to compare the experimental results to those of other learning systems. Solving problems of varying scales, different problems, and tracking changes in the domain generally requires domain-dependent set up, discovered through manual trial and error. A compelling direction for the future is to design benchmarks for domain-independent learning, making such comparisons possible and encouraging further work in this area.

In the future, a parallel implementation of DIAS should speed up and scale up problem-solving. It would be possible to run DIAS with larger search spaces in reasonable time. For high-dimensional domain-state and domain-action spaces, it may also be possible to fold the axes of the geo so that a single $(x, y)$ location can refer to more than one state or action in the domain space. This generalization, of course, would come at the expense of larger actor-action and actor-state space because each location would now have more than one value for domain state and action, but it could make it faster with high-dimensional domains. Another potential improvement is to design more actor types. While rule-set evolution performed well, it is a very general method, and it may be possible to design other methods that more rapidly and consistently adapt to specific problem domains as part of the DIAS framework.

\section{Conclusion}

DIAS is a domain-independent problem-solving system that can address problems with varying dimensionality and complexity, solve different problems with little or no hyperparameter tuning, and adapt to changes in the domain, thus implementing lifelong learning. These abilities are based on artificial-life principles, i.e.\ collective behavior of a population of actors in a spatially organized geo, which forms a domain-independent problem-solving medium. Experiments with DIAS demonstrate an advantage over a direct problem-solving approach, thus providing a promising foundation for scalable, general, and adaptive problem solving in the future.



\footnotesize
\bibliographystyle{apalike}


\end{document}